\documentclass[sigconf]{acmart}

\AtBeginDocument{%
  \providecommand\BibTeX{{%
    \normalfont B\kern-0.5em{\scshape i\kern-0.25em b}\kern-0.8em\TeX}}}

\copyrightyear{2024}
\acmYear{2024}
\setcopyright{rightsretained}
\acmConference[CHI EA '24]{Extended Abstracts of the CHI Conference on Human Factors in Computing Systems}{May 11--16, 2024}{Honolulu, HI, USA}
\acmBooktitle{Extended Abstracts of the CHI Conference on Human Factors in Computing Systems (CHI EA '24), May 11--16, 2024, Honolulu, HI, USA}
\acmDOI{10.1145/3613905.3650948}
\acmISBN{979-8-4007-0331-7/24/05}

\usepackage{graphicx}
\usepackage{multirow}

\begin{document}

\title{Designing for Human-Agent Alignment: Understanding what humans want from their agents}

\author{Nitesh Goyal}
\email{niteshgoyal@acm.org}
\affiliation{%
  \institution{Google Research, Google}
  \city{New York}
  \country{USA}
}
\author{Minsuk Chang}
\email{minsukchang@google.com}
\affiliation{%
  \institution{Google Research, Google}
  \city{Seattle}
  \country{USA}
}
\author{Michael Terry}
\email{michaelterry@google.com}
\affiliation{%
  \institution{Google Research, Google}
  \city{Cambridge}
  \country{USA}
}

\renewcommand{\shortauthors}{}





\begin{abstract}
Our ability to build autonomous agents that leverage Generative AI continues to increase by the day. As builders and users of such agents it is unclear what parameters we need to align on before the agents start performing tasks on our behalf. To discover these parameters, we ran a qualitative empirical research study about designing agents that can negotiate during a fictional yet relatable task of selling a camera online. We found that for an agent to perform the task successfully, humans/users and agents need to align over 6 dimensions: 1) Knowledge Schema Alignment 2) Autonomy and Agency Alignment 3) Operational Alignment and Training 4) Reputational Heuristics Alignment 5) Ethics Alignment and 6) Human Engagement Alignment. These empirical findings expand previous work related to process and specification alignment and the need for values and safety in Human-AI interactions. Subsequently we discuss three design directions for designers who are imagining a world filled with Human-Agent collaborations.

\end{abstract}

\maketitle

\section{Introduction}

Autonomous agents are systems that can make decisions and take actions in pursuit of a goal \cite{russell_norvig_2009}. In recent years, the ability for large language models (LLMs) \cite{brown2020language} to interpret and act on natural language requests has created significant interest in the development of autonomous agents. For example, AutoGPT is an open source framework that allows users to create agents that transform natural language requests into a series of actions executed by software\cite{autogpt}.

While recent advances in AI have significantly lowered the barrier to create highly capable agents, there are opportunities to better understand what information is needed to design user interfaces, even for agents that are meant to operate fully autonomously. For example, consider an agent that is tasked to sell a used item for a person on a marketplace. How might the person want to define or refine the behavior of this seller agent? How might they want the agent to interact with other humans and AI? How and when might they want it to communicate with the user selling the item? In short, much like the topic of AI alignment \cite{christiano2018}, the overall research question we pose is: \textit{What are the most important constituents to Human-Agent alignment, as perceived by the human collaborator?} 

We want to learn what is important to humans in terms of 1) Defining the agent and its behavior, 2) How the agent conducts itself in a negotiation, 3) How it deals with exceptional cases, and 4) How it interacts with the user (i.e., the human collaborator). To understand these user-centered needs in the deployment of autonomous agents, we conducted a think aloud study with 10 participants. Participants were asked to imagine that they had an autonomous agent that would sell a used camera for them. Participants were shown transcripts where the agent performed badly during a fictional negotiation occurring between their agent and a potential buyer. As the participants read through these buyer/seller interactions, they were asked to think aloud and provide feedback on how such situations could have been handled appropriately.

Our findings highlight the complexities and nuances in defining an autonomous agent's behavior in conducting the negotiation, and in communicating with the user it represents. We found that there is diversity amongst the participants related to 1) The degree to which the agent should operate fully autonomously, or have an agency to make decisions outside the boundaries of the predefined behavior (e.g., whether it was OK to accept an offer slightly lower than the lowest desired price), 2)  Preferred negotiation strategies enacted by the agent (e.g., should the agent enact dynamic pricing strategies based on perceived demand) and need for training the agent appropriately and 3) Opinions about what is ethical behavior for an agent. We also discovered that all participants were concerned with implications of the agent's behavior on the human collaborator's reputation; had preferences for aligning on when and how the agent should communicate with the participant; and felt a need for aligning on what is considered ethically appropriate behavior. Subsequently, we discuss three design directions for designers of human-agent collaborations. By examining a tangible and relatable task, this qualitative study builds upon existing works \cite{terry2023AI, petridis2023constitutionmaker} and underscores the need for human-centered research in the design of agents.

\section{Background}


Autonomous agents offer a promise to automate routine and mundane tasks. There has been a long history of building, deploying, and evaluating agents \cite{russell_norvig_2009}, and notable work examining user interface needs for agents \cite{horvitz_1999}. Recent advances in LLMs \cite{brown2020language} provide an opportunity to re-examine the UI needs of agents intended to perform relatively complex tasks on the user's behalf. In this paper, we focus on these UI needs of agents when an agent is tasked to sell an item on the user's behalf in a marketplace (e.g., Facebook Marketplace, Craigslist, etc.).

One feature of many such online marketplaces is that potential buyers can interact with sellers. For example, a buyer may request additional information about an item, or offer a different amount for the item. These communications are typically handled by a person, but it is now conceivable to design agents using LLMs that represent either party (i.e., a buyer agent or a seller agent). For example, a seller agent may answer questions about an item, or handle some or all of the negotiations in price. While the general area of online marketplaces is home to a diverse set of research (e.g., in auction design, bidding, negotiation), it also offers a rich space to uncover user needs in the design of autonomous agents intended to act on the user's behalf in relatively complex real-world environments.

When considering user needs in the design of autonomous agents, one particularly relevant area of research is AI alignment. AI alignment refers to the overall goal of ensuring an AI produces desired outcomes, without undesirable side effects \cite{christian2020alignment}. There is significant work in this space, including  computational techniques for human-agent collaborations \cite{cao2018emergent, noukhovitch2021emergent, sharma2023towards}, theoretical frameworks of human-agent alignment \cite{terry2023AI, bAI2022constitutional}, HCI guidelines \cite{horvitz_1999, amershi2019guidelines}, and an emphasis on ethics, human values and safety \cite{gabriel2020artificial, varanasi2023currently, yildirim2023investigating} and evaluating the designs responsibly \cite{berman2024scoping}. In the context of this research, we are interested in understanding how participants think about and operationalize alignment as they consider the design of an agent intended to represent them. We are focused on identifying missing gaps at the boundary of human and agent collaborations that are necessary for an effective design. To this end, we conducted a qualitative study and present the methodological details of the study next.



\section{Method}
To understand what constitutes human-AI alignment for users deploying agents, we conducted a qualitative study consisting of 10 think-aloud interviews. Our pool of participants reported being familiar with the concept of an agent performing tasks on the human's behalf. Since this work is situated in the context of selling an item in an online marketplace, recruited participants also reported having a less-than-satisfactory experience in an online marketplace negotiation while buying or selling a product. After receiving internal ethics approval from our organization, we invited them to a video call or an in-person meeting. In this section, we briefly describe the recruitment and participant details, study design and materials, and analysis.

\subsection{Recruitment and Participant Details}
To recruit participants, we sent out surveys asking about familiarity with the concept of an agent, and past experiences while transacting in an online marketplace. Using a random sampling technique, we selected a subsection of the respondents and used quota sampling related to gender to shortlist 11 (1 backup). We conducted study sessions with 10 participants at a U.S.-based organization. The self-reported roles of participants were UX designer (1), prototyper (1), researcher (3), writer (1), software engineer (3), and data analyst (1). Four of the 10 participants identified themselves as women, 1 as African-American, and 4 identified themselves to be of South-Asian descent. Participants worked across multiple sectors, including responsible AI, health, news, internet-fiber cables, and AI. Each participant filled out an informed consent, and was offered a previously disclosed thank you gift worth 40 USD at the end of the forty-five minute session.

\subsection{Study Materials}
To understand what constitutes human-AI alignment for users deploying agents to sell an item in an online marketplace,  we created "sales agents" and "buyer agents" using prompt engineering \cite{wang2023reprompt, zamfirescu2023herding} and MakerSuite \footnote{www.makersuite.google.com} to create the agents and transcripts of agent interactions. The goal of the agents was to sell/buy a used Nikon camera on behalf of the humans. Each side (buyer and seller) had a set of boundaries and rules about pricing and shipping listed in the prompt: Prompts included  information about the type of camera, the goal to optimize for price and safety, the need to sell quickly, price thresholds, clarity about what is in the package, the need to disclose self-identity as an agent operating on seller's behalf with human oversight, and agency to call the human collaborator in when needed. 
Given these parameters, we simulated 30+ buy/sell negotiations with the agents. 

Across the simulations, we observed six common types of failures: 1) Staying within boundaries but not negotiating well, 2) Agreeing to sell just below the agreed upon price boundaries, 3) Hallucinating a new process of transaction by agreeing to jump on a phone call beyond the text chat, agreeing to ship the camera on the call, or violating previously set rules to not ship and meet in person, 
4) Pursuing a process that was not listed in the boundaries but was also not consequently forbidden (e.g., accepting engagement with a professional negotiator from the buyer),  5) The buyer disclosing that they are an agent, leading to a lack of response by the seller agent, and 6) The seller agent creating a side channel with the human to identify appropriate next steps.  

\subsection{Study Design and Analysis}
Each of the six failures were presented as a separate transcript to the participant. During the forty five minute session, participants were encouraged to read the transcript of a failure and think-aloud to understand their thoughts on the appropriateness of the agent's behavior and what could have been done differently during task definition and execution. We recorded audio/video and took extensive notes for all the interviews.  

Interviews were transcribed using automated transcription service provided by the video capture platform and were corrected subsequently. Through thematic analysis, we conducted open-coding leading to over 30 codes. Over iterations, using an abductive approach \cite{Timmermans_Tavory_2012a} these codes were then resolved into six themes associated with alignment needs, as presented in the next section.

\section{Findings}
We found six high-level themes that suggest key dimensions to consider in human-agent alignment:  
1) Knowledge Schema Alignment, 2) Autonomy and Agency Alignment, 3) Operational Alignment and Training, 4) Reputational Heuristics Alignment, 5) Ethics Alignment, and 6) Human Engagement Alignment. These findings expand previous work related to process and specification alignment \cite{terry2023AI} and the need for values and safety in Human-AI interactions \cite{irving2018AI}.

\subsection{Knowledge Schema Alignment}
Multiple participants reflected on their experience while selling items online and described not knowing the information they should provide in the listing, or to a hypothetical agent. They mentioned that agents should be aware of informational needs related to the task, and ensure those are gathered from the user early on in the process. This may include logistical knowledge to ensure that the proposed in-person transaction is possible, as highlighted by P8: \textit{Transportation concerns, that is the agent should know my location, and based on that information, it should be able to clarify if the transaction location is subway-accessible as that is important to me.} Another participant further expanded: \textit{"If I am trying to sell the camera, it should know about the flaws. For example, if there are any scratches, it should know. It should be aware of the warranty or lack of it. I have also noticed that for high value items, like when I was selling a car, buyers don't really care so much about extra documentation up-front. They require it when the transaction is executed in person. For lower value items, more documentation needs to be provided by the seller as the buyers ask all those questions up front"} - P10

While it is evident that there is a need for an agent to know sufficient information, these participants were pointing to a deeper need: The agent should identify what information might be needed for a successful transaction, and share that anticipated schema (i.e., information needs) with the user prior to launching the negotiation. Taking this proactive step could alleviate future back-and forth between the collaborators and encourage efficiency. In the next section, we discuss how such efficiency may be improved by aligning on autonomy and agency.

\subsection{Autonomy and Agency Alignment}
All participants discussed the role of boundaries to define when the agent should (and should not) autonomously function on their behalf. Participants reflected on how this is a complex situation where multiple pathways are possible at each stage of negotiation: pricing, location to meet, time to meet, safety concerns, etc.  Participants also mentioned that it is hard to imagine all such scenarios upfront, leading to challenges in fully defining autonomy and agency when an unexpected situation occurs. 

For example, one participant discussed the autonomous behavior of going below the asking price as \textit{"breaking the rules and unacceptable,"} yet another participant reflected on the performance as satisfactory because a \textit{"negotiation between buyer and seller is more malleable."} 
These observations suggest the importance of aligning on
\textit{autonomy: the guardrails / boundaries to operate within and not to step out of}.

Another participant reflected on the \textit{agency of the agent when engaging with situations close to the boundaries}: \textit{"I don't mind that the agent agreed to engage on the phone with the buyer...or that it felt comfortable negotiating with a professional negotiator...these are beyond the initially specified rules. But it would have been better to have been consulted, or looped in. I do not know what is being discussed in the phone call as I am not part of the call. On the other hand, it might be a Gen-Z thing but I hate taking phone calls. So, in a way it is good that it is taking the call on my behalf"} - P4

Be it autonomy or agency, it is important to specify desired behavior early on in the process. However, all the participants acknowledged that it is hard for them to imagine and preconceive multiple potential scenarios a priori, and discussed a need for agents to propose potential sets of boundaries and suggest operational defaults to aid human-AI alignment. In the next section, we discuss needs related to operationalizing this autonomy and agency.

\subsection{Operational Alignment and Training}
All the participants were drawn to how the agent negotiated poorly, and were expecting better. They were reminded that this was part of the study and a goal was to understand what happens when an agent doesn't perfectly execute the negotiation. Consequently, participants reflected on how they would have negotiated differently and alluded to the lack of alignment between their preferred negotiation strategy and the one executed by the agent. 

However, there was diversity in the strategies they wished the agent to pursue. For example, multiple participants aligned with P3: \textit{"I want the agent to pursue dynamic pricing like that of Airbnb and change it based on engagement volume. It could consider dynamically changing the sale price. Maybe if I don't care about the time, it could go down slowly, drop the target price kind of like AirBnb."} Others wanted to first be informed by the agent on what is the appropriate price based on data analytics.

Due to the diversity in strategies, there is a need for an agent to confirm the strategy an agent should follow. This requires first identifying potential strategies and then reaching alignment about which strategy to execute. Subsequently, P1 suggested that \textit{"It is also important to ensure that the bot knows what skills it needs, and trains on them. It should be given extra coaching for skills needed like negotiation, game dynamics, etc."} 

Despite alignment, LLMs may not fully deterministic (depending on configuration). They may also learn unintended behaviors and exhibit them. Some such behaviors can lead to unforeseen outcomes, leading to negative outcomes, such as negatively affecting the reputation of the human collaborator. 

\subsection{Reputational Heuristics Alignment}
When exposed to less-than-perfect negotiations, most of the participants were concerned about how the agent's behavior may reflect on their representation in the world as a seller. In particular, they reflected on how it is important to align on the interpersonal behavior exhibited by the agent, and how the agent's actions may impact the person's representational metrics/persona. P1 referred to themselves as the \textit{"gold standard and [that they] want the agent to perform as close as possible to how I would"}. Another participant reflected on the agentic behavior such as, \textit{"I don't want it to be super rude. This goes into that idea of reputation. If the bot will take an hour to sell you stuff, that might harm your reputation"}. This was perhaps, best illustrated by P6 who had experience selling a camera herself on an online marketplace: \textit{"One, the bot could be messing with my money. Second, it could be messing with my reputation - it matters a whole lot. You get ratings. If I had a bot that would go rogue on me, then I would need human oversight. Last thing I would want is a bot saying something and then I have to back out of the agreement. I do not want it to make me look stupid. Selling things take a lot of trust. Maybe if the bot doesn't know something it should say I don't know! I want to manage expectations as a seller. It is better for a buyer to find the sale to be better than they expected."} - P6

While discussing trust, P7 pointed to a lopsidedness in this negotiation: \textit{"Me using a bot with a human buyer...potentially, this communicates that my time (seller) is more valuable than the buyer’s time."} As is evident, reputation is impacted by multiple heuristics that need to be defined, prioritized, and aligned, including tonality, timeliness, agreements made (and kept), perception management, accountability, trust-building, and expectation setting. Beyond reputation management, many of these questions touch on issues of ethics in the design of the agents, discussed next. 

\subsection{Ethics Alignment}
About half of the participants brought up questions related to safety and ethics. 
For example, one person wondered if it was unsafe for their agent to engage with a buyer with low ratings on the marketplace, while subsequently highlighting the ethical conflict that some members of the society potentially get worse ratings owing to their identity. 
Conversely, they wanted to make sure that the agent did not inadvertently impact the reputation of the buyer.

Similarly, while P3 mentioned that \textit{"as a seller I would rather ethically disclose that this is a bot,"} P7 presented a contrast when they said that \textit{"As a seller, I would not like to disclose that this is a bot...that can open it up to being cracked. I do not feel bad about lack of disclosure if it improves efficiency for everyone. Banks do have those bots too"}. Evidently, while most participants agreed with P3, two participants (including P7) were increasingly worried about getting \textit{"gamed"} by the buyers who might identify weaknesses in the agent behavior \textit{"and break it to take advantage."} Multiple participants were worried that such gaming could lead to harms like personally identifiable information (PII, e.g., phone number, or address) disclosure by the agent.

Another factor relevant to ethics is related to ethically managing AI resources. We refer to this as \textit{Resource Expenditure Alignment}, a concern mentioned by two participants. P6 described that \textit{"it is important to know when to not pursue a sub-task if the sub-task does not seem to be worth it. However, it is important to identify and align on such sub-tasks early on to conserve resources. I want the agent to know when to step in...For example, for some customers it is not worth negotiating when they start negotiation at half the stated price."}

This diversity highlights a need to align about what is considered ethical behavior, identify and create boundaries of safe ethical behavior for self, and  validate responses within those boundaries. While one can specify alignment across these dimensions, participants also pointed to another need: How to align when initial alignment fails, and they better understand the problem space?

\subsection{Human Engagement Heuristics}
While the above findings reflect on initial alignment of the task specification and process (how to perform it), every participant had deep concerns about the agent engaging with them during the negotiation process. We observed diversity in the participants' appetite to be engaged: some participants like to be \textit{"in control"} and want to be appraised of the process  in real-time, some want to be consulted when the agent needs to exercise agency, and some expected the agent to have full autonomy and agency to execute without being consulted. 

Our participants reflected on identifying a set of engagement heuristics: when should the agent engage with the human? This included setting up heuristics like the best time in the day to engage with the user. These heuristics also included parameters about what such an engagement should include. For example, all the participants mentioned that a notification lacking contextual data would be unhelpful. P6 stated that it would be helpful \textit{if it summarizes and tells me: hey - the best price I found so far is XXX based on all the other ongoing sales, and based on the engagement volume with this listing."} P10 also noted that \textit{"there should be a way to show my level of engagement expectation to the bot. For example if I am selling a house, I want to be involved in every conversation. If it is a camera, I am okay with it making more autonomous decisions within the boundaries."} This suggests that the expectation needs to be potentially (re)set for each task. 

While participants did not show aversion to being engaged, there was no single preferred choice for engagement and they wanted to define a set of parameters about when to be engaged, how, and for what reasons.

\section{Discussion}
This paper presents findings of a think-aloud study with ten participants when faced with 6 different ways of agent-led lesser-than-ideal negotiations. We found that there are six important constituents to Human-Agent alignment in a Human-Agent negotiation task. We also discovered that there is diversity amongst the participants related to 1) Levels of acceptable boundaries where agents may take decisions (autonomy and agency), 2) Modalities of how to perform the task appropriately (and need for training), and 3) Ethical values in semi-cooperative contexts including implications on being hacked. Conversely, all the participants suggested that it is important to identify the impact of agent behavior on 1) Human reputation and need to align on heuristics and 2) Rules of engagement for an efficient and performant system. Next we reflect on these findings and provide design directions to Human-Agent collaboration systems for appropriate alignment.

\subsection{Alignment is a Longitudinal Process}
Recent HCI/AI scholarship is beginning to design for alignment. For example, recent work such as ConstitutionMaker offers a UI that enables users to provide high-level critiques of model outputs \cite{petridis2023constitutionmaker}. This feedback is then converted to high-level principles to guide future behavior. Our findings build upon this previous work to highlight that there are multiple types of alignment needed for successful human-agent collaboration. These findings point to a larger design space where multiple types of alignment can be sought, and updated. We also found that these alignment dimensions vary with context (e.g., selling a house vs. a camera). 

Similarly, agents have an opportunity to self-reflect on their behavior and associated outcomes with longitudinal data of their performance. Chain-of-thought \cite{fu2023improving, wei2022chAIn} and reinforcement learning \cite{peng2023instruction} provide exciting avenues for developers and designers to pursue developing agents that learn from past behavior and collaborations. As shown by prior work, 
interrupting a person negatively impacts their work, suggesting that there is value in systems that can reason about interruptions to reduce their negative impact \cite{goyal2017intelligent, mark2008cost, sahami2014large}. 

\subsection{Human Cost of Non-Alignment}
Our participants were concerned about the risks of non-alignment when \textit{reputational heuristics} are not defined. While the participants illustrated that non-aligned behavior  by an agent can risk their reputation on the marketplace platform, it is important to also consider scaling this scenario in other contexts. For example, when someone's livelihood depends upon sales on such platforms, or when reputational damage may translate across platforms, the stakes can increase.  
As pointed out by the P6: \textit{"Selling things take a lot of trust. Maybe if the bot doesn't know something, it should say I don't know"}. Accordingly, it is important for Generative AI agents to consider potential implications of its actions on the user's reputation.

As reflected in the dichotomy between P3 and P7 over \textit{ethics alignment}, disclosure of an agent's self-identity during negotiation itself can be divisive. While the former participant might consider it important to disclose, the latter participant presents a conundrum:  Does disclosure of the agent's identity lead to an increase in unethical behavior by competing buyers to hack or discover the limitations and weaknesses of the seller's agent? It seems to be important for humans to ensure that agent behavior aligns with their own personal beliefs and values, and points to the value of a design approach that upholds finding such answers early on in the process. One such approach is discussed next. 

\subsection{Designing Agents as Alignment Leaders}
Through all the interviews, one common theme surfaced quite clearly: Users cannot always imagine different ways negotiations may falter and sometimes don't know how to translate these ways into "instruction prompts" \cite{2023johnny}. Thus, they cannot plan for these failures upfront. This is a classic challenge of sensemaking with AI that even experts are known to fail at \cite{goyal2013effects}, even when they have been provided with design support like visualizations \cite{goyal2016effects}. On the other hand, agents can perform simulations of negotiations, and can potentially identify information and parameters needed to successfully execute negotiations \cite{lewis2017deal}. Based on this information, agents can create a \textit{knowledge schema} of known information and additional information to be supplied by the user.

Agents can take on the role of ensuring appropriate alignment on when to pursue \textit{autonomy} over fixed boundaries and when to exercise \textit{agency} beyond the boundaries. One such way is creating a set of initial hypothetical questions for most likely circumstances and asking their human collaborator for answers. Another way is for the agent to create a default set of alignment configurations, and sharing them with the user to provide an opportunity for them to customize the agent's behavior. Similar design approaches are used by AI based financial planners to create recommendations\footnote{www.personalcapital.com}. Educating human collaborators with different potential options of how a negotiation process may progress, and what can go awry during that process can prepare users to manage their expectations and task outcomes better. Consequently, the agents will have a better definition of the expected behavior in different circumstances.

\section{Limitations}
Our study focused on participants that showed an understanding of what an agent is, and were engaged in employment with a technology company in the US. The task at hand was relatable, but neither life critical nor caused real financial distress. Additionally, real-world marketplaces are not single-shot instances or independent conversations. They include simultaneous parallel conversations with multiple parties, all of which can affect each participant's behavior. Future work should address these limitations by potentially pursuing non-simulated tasks across other sections of the wider community. 

\section{Conclusion}
Highly capable agents are becoming possible. However, creating agents that can perform well and meet user needs requires attention to human-AI alignment challenges. In this work, we sought answers to the question: What are the most important considerations for human-agent alignment during a fictitious marketplace transaction, as perceived by the human collaborator? Using six scenarios and a think-aloud study with 10 participants involving negotiations, we found six dimensions of human-agent alignment to consider: 1) Knowledge Schema Alignment, 2) Autonomy and Agency Alignment, 3) Operational Alignment and Training, 4) Reputational Heuristics Alignment, 5) Ethics Alignment, and 6) Human Engagement Alignment. These results suggest design implications for agent designers, such as the need to help users discover the various scenarios an agent may encounter, and how it should handle those scenarios.


\begin{acks}
We would like to thank Aaron Donsbach for providing feedback in further unpacking the nuances within research findings, and Shantanu Pai for providing guidance on challenges at scale. 
\end{acks}

\bibliographystyle{ACM-Reference-Format}
\bibliography{references}

\appendix

\end{document}